\documentclass[10pt, a4paper]{article}

\usepackage{lrec-coling2024} 

\title{A Single Linear Layer Yields Task-Adapted Low-Rank Matrices}

\name{Hwichan Kim$^{\dagger \ast}$\thanks{$^\ast$This work was carried out during his CyberAgent internship period.}, Shota Sasaki$^\ddag$, Sho Hoshino$^\ddag$, Ukyo Honda$^\ddag$} 

\address{$\dagger$Tokyo Metropolitan University, $\ddag$CyberAgent \\
         $\dagger \ddag$ Tokyo, Japan \\
         kim-hwichan@ed.tmu.ac.jp, \{sasaki\_shota, hoshino\_sho, honda\_ukyo\}@cyberagent.co.jp \\}

\abstract{
Low-Rank Adaptation (LoRA) is a widely used Parameter-Efficient Fine-Tuning (PEFT) method that updates an initial weight matrix $W_0$ with a delta matrix $\Delta W$ consisted by two low-rank matrices $A$ and $B$.
A previous study suggested that there is correlation between $W_0$ and $\Delta W$.
In this study, we aim to delve deeper into relationships between $W_0$ and low-rank matrices $A$ and $B$ to further comprehend the behavior of LoRA.
In particular, we analyze a conversion matrix that transform $W_0$ into low-rank matrices, which encapsulates information about the relationships. 
Our analysis reveals that the conversion matrices are similar across each layer. 
Inspired by these findings, we hypothesize that a single linear layer, which takes each layer’s $W_0$ as input, can yield task-adapted low-rank matrices. 
To confirm this hypothesis, we devise a method named Conditionally Parameterized LoRA (CondLoRA) that updates initial weight matrices with low-rank matrices derived from a single linear layer. 
Our empirical results show that CondLoRA maintains a performance on par with LoRA, despite the fact that the trainable parameters of CondLoRA are fewer than those of LoRA. 
Therefore, we conclude that "a single linear layer yields task-adapted low-rank matrices." 
 \\ \newline \Keywords{Pretrained Language Model, Parameter-Efficient Fine-tuning, Low-Rank Adaptation}}

\begin{document}

\maketitleabstract

\section{Introduction}
In natural language processing (NLP) area, it is common practice to fine-tune pre-trained language models (PLMs) \cite{devlin-etal-2019-bert,lewis-etal-2020-bart,NEURIPS2020_1457c0d6} using task-specific data.
As the scale of these PLMs has grown considerably, the computational resources required for fine-tuning all parameters have escalated, presenting a substantial computational challenge.
In recent years, parameter-efficient fine-Tuning (PEFT) methods, which use a limited number of additional parameters, have been proposed to address this issue.
PEFT methods include prompt-tuning \cite{lester-etal-2021-power}, prefix-tuning \cite{li-liang-2021-prefix}, and low-rank adaptation (LoRA) \cite{hu2022lora}, etc.
These methods reduce computational costs to fine-tune PLMs while achieving comparable performance to fine-tuning all of the parameters.

Among the PEFT methods, LoRA has been prominent in NLP area because it shows stable and good performance across various NLP tasks and PLMs \cite{pu2023empirical}. 
LoRA fixes an initial weight matrix $W_0$ and updates $W_0$ with a delta matrix $\Delta W$ consisting of trainable low-rank matrices $A$ and $B$, significantly reducing the number of trainable parameters compared to fine-tuning all parameters.
Subsequent studies \cite{zhang2023adaptive,valipour-etal-2023-dylora}  have analyzed several aspects of LoRA to achieve potentially more effective and efficient PLM fine-tuning.
\citet{hu2022lora} performed an analysis of the relationship between $W_0$ and trained $\Delta W$ ($= BA$), and they revealed that there is a correlation between $W_0$ and $\Delta W$.
This finding implies the existence of certain relationships between the initial weight matrix $W_0$ and the low-rank matrices $A$ and $B$.

\begin{figure*}[t]
  \begin{minipage}[b]{0.33\hsize}
    \centering
    \includegraphics[scale=0.4]{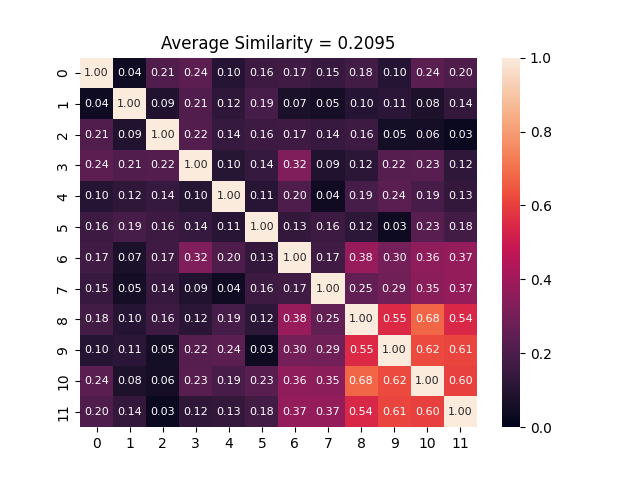}
    \subcaption{$W^{\mathrm{value}, l}_{0 \to A}$}\label{figure:va_sim}
  \end{minipage}
  \begin{minipage}[b]{0.33\hsize}
    \centering
    \includegraphics[scale=0.4]{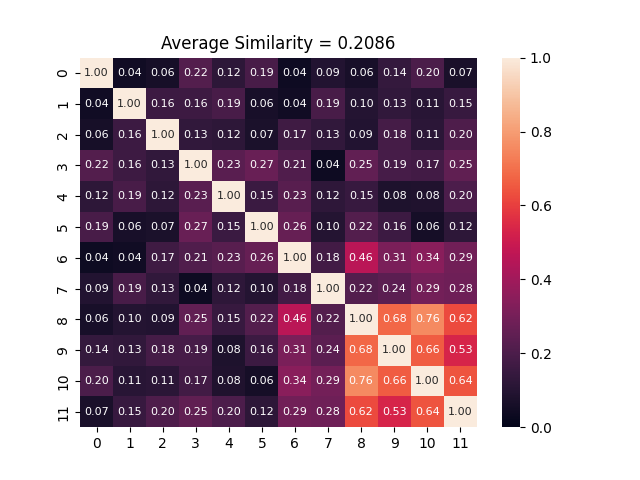}
    \subcaption{$W^{\mathrm{value}, l}_{0 \to B}$}\label{figure:vb_sim}
  \end{minipage}
  \begin{minipage}[b]{0.33\hsize}
    \centering
    \includegraphics[scale=0.4]{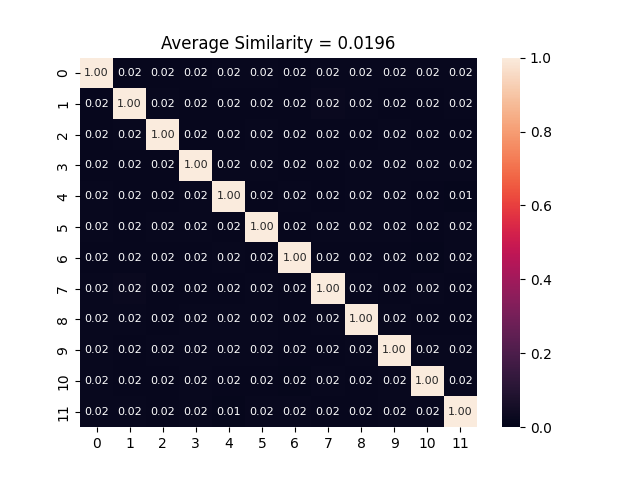}
    \subcaption{Random matrix}\label{figure:random_sim}
  \end{minipage}
  \caption{Normalized subspace similarities between each layer's conversion matrices and random matrices. Average similarity refers to the average of elements excluding the diagonal elements.}\label{figure:subspace_sim}
\end{figure*}



In this study, we conduct an in-depth analysis of the relationships between initial weight matrices $W_0$ and low-rank matrices $A$ and $B$ to gain a deeper understanding of LoRA behavior.
Specifically, we analyze a conversion matrix that transforms $W_0$ into $A$ or $B$ under the assumption that it roughly represents their relationships.
Our analysis shows that similarities between each layer's conversion matrix are very high.
This empirical observation implies a commonality in the relationships between the initial weight matrices and low-rank matrices regardless of layers.
Inspired by the results, we hypothesize that a single linear layer, which takes each layer's $W_0$ as input, can produce task-adapted low-rank matrices of each layers.

To confirm our hypothesis, we design a method named Conditionally Parameterized LoRA (CondLoRA) that fine-tune PLMs with low-rank matrices derived from a single linear layer (Figure \ref{figure:condlora}).
Our experiments demonstrate that CondLoRA achieves competitive performance compared to LoRA in GLUE tasks.
Notably, CondLoRA can reduce the number of trainable parameters compared to LoRA, because its parameters are constant regardless of target layers.
The success of CondLoRA suggests potential avenues for further minimization
of trainable parameters in LoRA variants.
Our contributions in this study are twofold: 
\begin{enumerate}
\item We reveal that conversion matrices that transform initial weight matrices into trained low-rank matrices are similar across each layer, which means that there is similar relationship regardless of layers.
\item We demonstrate that CondLoRA  achieves performance comparable to the already parameter-efficient LoRA with fewer parameters. This outcome suggests that task-adapted low-rank matrices can be yielded by a single linear operation.\footnote{We will publish the code used in our experiments.}
\end{enumerate}

\begin{table}[tp]
\begin{center}
\scalebox{1.0}{
\begin{tabular}{ll}
\toprule
Hyperparameters & Value \\
\midrule
Batch Size & 16 \\
Optimizer & Adam \\
Scheduler & Linear \\
Target Modules & \{query, value\} \\
Target Layers & \{1, 2, ..., 12\} \\
LoRA $r$ & 8 \\
LoRA $\alpha$ & 8 \\
Max Seq. Len. & 512 \\
\bottomrule
\end{tabular}
}
\end{center}
\caption{Hyperparameters used in our experiments.}
\label{table:hyperparameter}
\end{table}

\section{Preliminaries for LoRA}


A diff-planing method \cite{guo-etal-2021-parameter} updates an initial weight matrix $W^{m, l}_0 \in \mathbb{R}^{d_1\times d_2}$ using an trainable matrix $\Delta W^{m, l} \in \mathbb{R}^{d_1\times d_2}$.
Where $m \in \{m_1, ..., m_k\}$ and $l \in \{1, 2, ..., N\}$ are target module (e.g., query, value, etc.) and layer, respectively, and $N$ is a total number of layers.
\citet{hu2022lora} proposed a PEFT method named LoRA, which decompose $\Delta W^{m, l}$ into two low-rank weight matrices:

\begin{equation}
\label{equation:lora}
W^{m,l}_0 + \Delta W^{m,l} = W^{m,l}_0 + B^{m,l}A^{m,l}
\end{equation}

\noindent where $A^{m,l} \in \mathbb{R}^{r \times d_1}$ and $B^{m,l} \in \mathbb{R}^{d_2 \times r}$ with $r \ll {d_2, d_1}$.
$A^{m,l}$ and $B^{m,l}$ are trained by downstream-task data. 
Their experiments demonstrated that LoRA achieves comparable or even better performance than full fine-tuning while reducing the numbers of trainable parameters.

In addition, they analyzed several aspects of trained $A^{m,l}$, $B^{m,l}$, and ${\Delta W^{m,l}}$ using normalized subspace similarity.
They defined normalized subspace similarity $\phi(\cdot)$ between matrices $X \in \mathbb{R}^{d^{X}_1 \times d^{X}_2}$ and $Y \in \mathbb{R}^{d^{Y}_1 \times d^{Y}_2}$ as:

\begin{equation}
\label{equation:subspace_sim}
\phi(X, Y, i, j) = \frac{\| U^{i\top}_X U^{j}_Y \|^2_\mathrm{F}}{\mathrm{min}(i,j)} \in [0, 1]
\end{equation}

\begin{table*}[tp]
\begin{center}
\scalebox{1.0}{
\begin{tabular}{lccccccccc}
\toprule
 & MNLI & SST-2 & MRPC & CoLA & QNLI & QQP & RTE & STS-B & Avg. \\
\midrule
LoRA & 86.6 & 93.7& 86.2 & 61.2 & 92.0 & 90.5 & 74.3 & 89.3 &  83.38 \\
CondLoRA & 86.5 & 93.8 & 86.6 & 61.1 & 91.8 & 90.1 & 74.2 & 90.3 & 83.42 \\
$\Delta$ & -0.1 & 0.1 & 0.4 & -0.1 & -0.2 & -0.4 & -0.1 & 1.0 & 0.04 \\
\bottomrule
\end{tabular}
}
\end{center}
\caption{Evaluation on GLUE tasks. These scores are the average of three models.}
\label{table:gleu}
\end{table*}

\noindent where $U_X$ is a left or right unitary matrix
and $U^i_X$ is top-$i$ singular vectors of $U_X$. Therefore, when $\phi(X, Y, i, j)$ is close to $1$, it means the singular vector spaces between $X$ and $Y$ are similar. \citet{hu2022lora} measured subspace similarities between each $W^{m, l}_0$ and $\Delta W^{m,l}$, and showed that the similarities are higher than those of random Gaussian matrices.
This result suggests there are relationships between an initial weight matrix $W^{m,l}_0$ and low-rank matrices $A^{m, l}$ and $B^{m, l}$.

\section{Commonality of Relationships across Layers}
\label{section:rq1}
In this study, we conduct an in-depth analysis of the relationships between initial weight matrices $W_0$ and low-rank matrices $A$ and $B$ to comprehend LoRA behavior.
To analyze the relationships, we use a conversion matrix that transform $W^{m, l}_0$ to $A^{m, l}$ or $B^{m, l}$, under the assumption that it roughly represents their relationships.
The analyses of the conversion matrix are expected to provide a deeper understanding of the relationship.

Conversion matrices $W^{m,l}_{0 \to A}$ and $W^{m,l}_{0 \to B}$ satisfy $W^{m, l}_0 W^{m,l}_{0 \to A} = (A^{m, l})^\top$ and $W^{m, l}_0 W^{m,l}_{0 \to B} = B^{m, l}$, respectively.\footnote{Notably, we assume that $d_1$ and $d_2$ are the same value in our experiments, because we used RoBERTa base and target modules are query and value. If $d_1$ and $d_2$ are different values, it is necessary to apply some extensions for calculating a conversion matrix.} $(A^{m, l})^\top$ is a transposed matrix of $A^{m, l}$. Therefore, the conversion matrices are:

\begin{equation}
\label{equation:w_to_a}
W^{m,l}_{0 \to A} = (W^{m,l}_0)^{-1} (A^{m, l})^\top  \in \mathbb{R}^{d_2 \times r}
\end{equation}

\begin{equation}
\label{equation:w_to_b}
W^{m,l}_{0 \to B} = (W^{m,l}_0)^{-1} B^{m, l}  \in \mathbb{R}^{d_2 \times r}
\end{equation}

\noindent where $(W^{m,l}_0)^{-1}$ is an inverse matrix of $W^{m,l}_0$.

In this study, we investigate the similarity of conversion matrices across layers.
Specifically, we measure normalized subspace similarities (Equation \ref{equation:subspace_sim}) between conversion matrices of each layer.
When the similarities are high, it suggests there is a similar relationship between initial weight matrices $W^{m,l}_0$ and low-rank matrices $A^{m, l}$ and $B^{m,l}$ across layers.

\subsection{Experimental Settings}
\label{section:experiment1}
We used RoBERTa base \cite{DBLP:journals/corr/abs-1907-11692} as a base model and HuggingFace Transformers \citeplanguageresource{wolf-etal-2020-transformers}.\footnote{\url{https://huggingface.co/roberta-base}} 
We used PEFT library\footnote{\url{https://github.com/huggingface/peft}} and a single NVIDIA A100 40GB for LoRA tuning.
We fine-tuned the model using GLUE \citeplanguageresource{wang-etal-2018-glue} dataset.
We set the hyperparameters except for learning rates following \citet{hu2022lora} as shown in Table~\ref{table:hyperparameter}.
We fine-tuned the models using only 90\% of training set.
We allocated the remaining 10\% for development and used the official GLUE development set as our test data because GLUE dataset does not provide an official test set.
We set max epochs to 50 in MNLI and QQP and 100 in other tasks, respectively.
Based on evaluation scores in development data, we searched learning rates through Optuna \cite{optuna_2019}\footnote{\url{https://optuna.org/}} and selected the best checkpoint.
For evaluation metrics, we used the overall (matched and mismatched) accuracy for MNLI, Matthew’s correlation \cite{MATTHEWS1975442} for CoLA, Pearson correlation for STS-B, and accuracy for other tasks.
To measure the normalized subspace similarity (Equation \ref{equation:subspace_sim}), we used a left unitary matrix and set $i$ and $j$ to be 8 ($= r$).

\begin{figure}[t]
 \centering
 \includegraphics[scale=0.45]{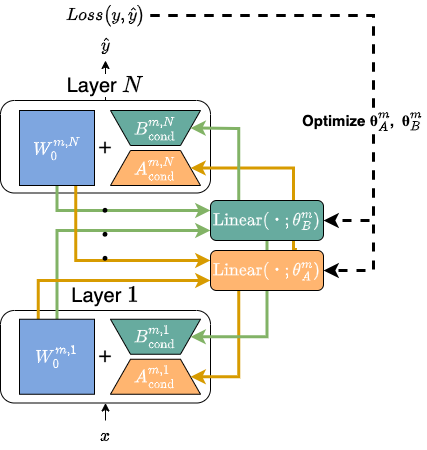}
 \caption{An overview of CondLoRA.}
 \label{figure:condlora}
\end{figure}

\subsection{Experimental Results}
Figure \ref{figure:subspace_sim} shows normalized subspace similarities between conversion matrices (Equations \ref{equation:w_to_a} and \ref{equation:w_to_b}) of each layer and those of random Gaussian matrices.
Due to the limited number of pages, we only show the similarities of conversion matrices from a model fine-tuned by MNLI dataset and value modules.
The similarities of conversion matrices were higher than those of random matrices.
This result implies a commonality in the relationships between the initial weight matrices $W^{m, l}_0$ and low-rank matrices $A^{m, l}$ and $B^{m, l}$ regardless of layers.
Inspired by this result, we hypothesize that a single linear layer, which takes each layer’s $W^{m,l}_0$ as input, can produce task-adapted low-rank matrices $A^{m, l}$ (or $B^{m, l}$) of each layer. 

In addition, this analysis reveals another noteworthy observation that the similarities between the deeper layers are extremely high. 
We would like to investigate the underlying causes in future work (See Section~\ref{section:limitations} for details).

\begin{table*}[tp]
\begin{center}
\scalebox{0.91}{
\begin{tabular}{llcccccccccccc}
\toprule
& & \multicolumn{12}{c}{$l$-th layer} \\
$X$ & $Y$ & 1 & 2 & 3 & 4 & 5 & 6 & 7 & 8 & 9 & 10 & 11 & 12 \\
\midrule
 $A^{\mathrm{value},l}$ & $A^{\mathrm{value}, l}_\mathrm{cond}$  & 0.05 & 0.05 & 0.06 & 0.04 & 0.04 & 0.08 & 0.03 & 0.04 & 0.03 & 0.06 & 0.05 & 0.07 \\
 $B^{\mathrm{value},l}$ & $B^{\mathrm{value},l}_\mathrm{cond}$  & 0.04 & 0.03 & 0.03 & 0.05 & 0.05 & 0.04 & 0.04 & 0.03 & 0.02 & 0.07 & 0.08 & 0.10 \\
 $\Delta W^{\mathrm{value},l}$ & $\Delta W^{\mathrm{value},l}_\mathrm{cond}$  & 0.04 & 0.03 & 0.03 & 0.05 & 0.05 & 0.05 & 0.04 & 0.03 & 0.02 & 0.07 & 0.09 & 0.12 \\
\bottomrule
\end{tabular}
}
\end{center}
\caption{Normalized subspace similarity between matrices from LoRA and CondLoRA.}
\label{table:lora_condlora_sim}
\end{table*}

\begin{table}[tp]
\begin{center}
\scalebox{0.95}{
\begin{tabular}{lcc}
\toprule
 & \begin{tabular}{c} Trainable \\ parameters \end{tabular} & \begin{tabular}{c} Speed  \\ (examples/s) \end{tabular} \\
\midrule
LoRA & 294,912 &  39.652 \\
CondLoRA & 24,576  & 40.303 \\
\bottomrule
\end{tabular}
}
\end{center}
\caption{Trainable parameters and speed during training.}
\label{table:parameter}
\end{table}

\section{Can a Single Linear Layer Yield Task-Adapted Low-Rank Matrices?}
\label{section:rq2}

In this section, to confirm our hypothesis (Section \ref{section:rq1}), we design a method named Conditionally Parameterized LoRA (CondLoRA) that fine-tune PLMs with low-rank matrices derived from a single linear layer.
CondLoRA finds low-rank matrices $A^{m, l}_{\mathrm{cond}}$ and $B^{m, l}_{\mathrm{cond}}$ using linear layers as follows:

\begin{equation}
\label{equation:condlora_a}
A^{m, l}_\mathrm{cond}= \mathrm{Linear}(W^{m, l}_0; \theta^m_A)^\top \in \mathbb{R}^{r \times d_1}
\end{equation}

\begin{equation}
\label{equation:condlora_b}
B^{m, l}_\mathrm{cond} = \mathrm{Linear}((W^{m, l}_0)^\top; \theta^m_B) \in \mathbb{R}^{d_2 \times r}
\end{equation}

\begin{equation}
\label{equation:condlora_delta}
\Delta W^{m, l}_\mathrm{cond} = B^{m, l}_\mathrm{cond} A^{m, l}_\mathrm{cond}
\end{equation}

\noindent where $\theta^m_A \in \mathbb{R}^{d_2 \times r}$ and $\theta^m_B \in \mathbb{R}^{d_1 \times r}$ are trainable parameters. CondLoRA train $\theta^{m}_A$ and $\theta^{m}_B$ using downstream task data. We provide an overview of CondLoRA as shown in Figure \ref{figure:condlora}.

One of the advantages of CondLoRA is its ability to decrease the numbers of trainable parameters.
LoRA requires $(d_1 \times r + d_2 \times r) \times k \times N$ trainable parameters.
However, CondLoRA requires $(d_1 \times r + d_2 \times r) \times k$ trainable parameters regardless of $N$, because it use a linear layer per target modules and low-rank matrices.
To substantiate our hypothesis, we conduct a comparative analysis of LoRA and CondLoRA based on their performance in GLUE tasks.

\subsection{Experimental Results}

Table \ref{table:gleu} shows the evaluation scores of validation data in each task.\footnote{We used the same settings in Subsection \ref{section:experiment1}.} 
The average scores (Avg.) across all the tasks are nearly equal between LoRA and CondLoRA.
Furthermore, CondLoRA outperforms LoRA in SST-2, MRPC, and STS-B tasks.
We also performed a pairwise $t$-Test to measure the statistical significance of the performance difference. 
The $p$-values were over 0.01 in all the tasks, indicating that CondLoRA achieves competitive performance compared to LoRA.
From the experimental results, we conclude that ``a single linear layer yields  task-adapted low-rank matrices''.

\subsection{Analysis}

\paragraph{The numbers of trainable parameters.}
As explained at Section \ref{section:rq2}, the numbers of trainable parameters of CondLoRA is constant regardless of the number of target layers.
We show the numbers of trainable parameters of CondLoRA and LoRA in Table \ref{table:parameter}.
Table \ref{table:parameter} shows that CondLoRA reduces the numbers of trainable parameters to $\frac{1}{12}$ compared to LoRA, because RoBERTa base is consisted by 12 layers and we targeted all layers.

\paragraph{Speed.}
CondLoRA has extra calculations compared to LoRA, because it determines $A^{m, l}_\mathrm{cond}$ and $B^{m, l}_\mathrm{cond}$ based on $W^{m, l}_0$ (Equations \ref{equation:condlora_a} and \ref{equation:condlora_b}).
There is no difference in inference speed between LoRA and CondLoRA, since the calculations are performed only once when loading a model.
However, during training, CondLoRA may takes extra time compared to LoRA because the calculations are required per each iteration.
Therefore, we quantified speeds, the numbers of examples processed per second, during training of both LoRA and CondLoRA as shown in Table \ref{table:parameter}.
Contrary to expectations, CondLoRA is slightly faster than LoRA.
We consider that the delay from the calculations for low-rank matrices are offset by the backpropagation process, because the trainable parameters (i.e. the parameters to be updated by backpropagation) are fewer than LoRA.

\paragraph{Similarity between low-rank matrices.}
Finally, we measured normalized subspace similarity between $A^{\mathrm{value}, l}$, $B^{\mathrm{value}, l}$ and $A^{\mathrm{value}, l}_\mathrm{cond}$, $B^{\mathrm{value}, l}_\mathrm{cond}$, respectively.
We used right and left unitary matrices for $A$ and $B$, respectively, and set 8 as $i$ and $j$.
Table \ref{table:lora_condlora_sim} demonstrates that while the similarities are not exceedingly high, they are higher than those of random Gaussian matrices.\footnote{The similarities between random matrices are less than 0.01.} 
A similar trend was also observed in  query modules. 
This result implies that LoRA and CondLoRA, to some degree, obtain similar low-rank matrices.

\section{Conclusion and Future Work}
In this study, we demonstrated that similar relationships exist between the initial weight matrices and low-rank matrices regardless of layers. 
Inspired by this analysis, we designed CondLoRA, which updates initial weights with low-rank matrices derived from a single linear layer. 
We showed that CondLoRA achieves competitive performance compared to LoRA, while trainable parameters are reduced.

Although out of scope of this study, we believe that CondLoRA has the potential to reduce the trainable parameters of other LoRA variants such as AdaLoRA \cite{zhang2023adaptive}.
Therefore, for future work, we would like to apply CondLoRA to other LoRA variants and investigate its effectiveness (See Section~\ref{section:limitations} for details).

\label{sec:append-how-prod}

\section*{Limitations}
\label{section:limitations}
Although our analyses provided novel insights to achieve more effective and efficient PLMs fine-tuning, our current work has the following limitations.

\paragraph{Analyses in other models and tasks.}
We used  RoBERTa base and GLUE tasks to conduct the analyses of Sections \ref{section:rq1} and \ref{section:rq2}.
It is not immediately clear whether these conclusions would hold true for other PLMs and tasks. 
Therefore, for future work, we are interested in analyzing other PLMs and various tasks to verify if similar results can be achieved irrespective of PLMs or tasks.

\paragraph{Analyses of conversion matrices}
In this study, we have investigated whether conversion matrices are similar across each layer. 
We also observed that the similarities between the deeper layers are extremely high. 
\citet{phang-etal-2021-fine} fine-tuned all of the PLM's parameters with task-specific data and measured similarities between each layer.
They showed that the similarities between the deeper layers are high compared to others.
We would like to investigate that a causal relationship between layer similarities of the model fine-tuned all of the parameters and those of conversion matrices.
In addition, we have not conducted an analysis of the conversion matrix itself. 
More detailed analyses about the conversion matrix will provide further insight into the nature of these relationships.

Additionally, Equation \ref{equation:w_to_a} and \ref{equation:w_to_b} are not be able to use to matrices where $d_1$ and $d_2$ are different, as inverse matrices are not be able to find for such matrices. 
Therefore, in order to conduct analyses using conversion matrices for other modules, such as feed-forward layers, we aim to devise a more generalized method for finding conversion matrices in future work.

\paragraph{Evaluation of CondLoRA with LoRA variants}
CondLoRA, a method that finds low-rank matrices for each layer using a linear layer, can also be applied to other LoRA variants. 
For instance, CondLoRA could be applied to AdaLoRA \cite{zhang2023adaptive}, decomposing $\Delta W^{m, l}$ into the form of singular value decomposition $P^{m,l}\Lambda^{m,l}Q^{m,l}$, using trainable parameters $\theta^{m}_P$, $\theta^{m}_\Lambda$, and $\theta^{m}_Q$ for each matrix. 
Therefore, further investigation into the effectiveness of CondLoRA when applied to other LoRA variants remains a challenge for future research.

\section*{Ethical Consideration}
In recent years, bias in data has become an issue.
Training a models, not just on training using CondLoRA, on biased datasets can result in unjustified predictions or the generation of pejorative content directed towards specific individuals or groups.
Intuitively, if the parameters are abundant, the effect of bias will be distributed across each parameter, but if they are small, all parameters may be affected by bias.
Since CondLoRA has very small trainable parameters compared to other fine-tuning methods (such as full parameter tuning and the other PEFT methods), it may be more susceptible to the effects of bias.
Therefore, when using CondLoRA, sufficient attention should be paid to the problem of bias.

\section*{Acknowledgments}
We would like to thank Tosho Hirasawa for his help with proposing CondLoRA and writing our paper.

\section{Bibliographical References}\label{sec:reference}

\bibliographystyle{lrec-coling2024-natbib}
\bibliography{lrec-coling2024-example}

\section{Language Resource References}
\label{lr:ref}
\bibliographystylelanguageresource{lrec-coling2024-natbib}
\bibliographylanguageresource{languageresource}

\end{document}